# Blind Analysis of CT Image Noise Using Residual Denoised Images

Sohini Roychowdhury, *Member, IEEE,* Nathan Hollcraft, Adam M. Alessio, *Senior Member, IEEE*

*Abstract–* CT protocol design and quality control would benefit from automated tools to estimate the quality of generated CT images. These tools could be used to identify erroneous CT acquisitions or refine protocols to achieve certain signal to noise characteristics. This paper investigates blind estimation methods to determine global signal strength and noise levels in chest CT images. Methods: We propose novel performance metrics corresponding to the accuracy of noise and signal estimation. We implement and evaluate the noise estimation performance of six spatial- and frequency- based methods, derived from conventional image filtering algorithms. Algorithms were tested on patient data sets from whole-body repeat CT acquisitions performed with a higher and lower dose technique over the same scan region. Results: The proposed performance metrics can evaluate the relative tradeoff of filter parameters and noise estimation performance. The proposed automated methods tend to underestimate CT image noise at low-flux levels. Initial application of methodology suggests that anisotropic diffusion and Wavelet-transform based filters provide optimal estimates of noise. Furthermore, methodology does not provide accurate estimates of absolute noise levels, but can provide estimates of relative change and/or trends in noise levels.

## I. Introduction

Computed tomography (CT) is a widely used imaging technology that allows volumetric visualization of the internal x-ray attenuation of a scanned object [1]. The CT imaging community faces concerns regarding the risks of diagnostic radiation exposure. This reality encourages the implementation of optimized CT imaging protocols with sufficient diagnostic *image quality* at the lowest achievable dosage levels. The goal of this paper is to analyze various parametric image denoising algorithms for estimating signal enhancement and noise estimation metrics for automated image quality estimation applications.

In brief, this work seeks to perform blind estimation of signal strength (feature contrast, resolution) from optimally denoised images, and estimate noise levels from the residual image obtained by subtracting the denoised image from the original CT image. In this paper, we focus on the estimation of image noise. This will be performed from evaluating the residual image obtained from difference of the original image and a filtered image. The intuition of this approach derives from the logic that in some imaging scenarios, all of the image noise could be contained in certain frequency bands and all of the signal in different bands. A band-pass filter could be used separate the signal from the noise (i.e., the residual image could be the noise image). In CT images and most practical applications, noise and signal have overlapping frequency content are therefore it is difficult to separate and estimate them without prior knowledge. Our work seeks to explore the ability to blindly separate signal and noise in CT images.

There are a wealth of potential approaches for image filtration in CT images including weighted averaging of neighborhood pixels [2], bilateral filters [2], patch-wise non-local means (PWNLP) based on a sparsity prior of a learned dictionary [3]. Wavelet-based smoothing methods [4], and Fourier-domain error based (FDE) denoising approaches [5]. This methodology could be developed with testing in simulations where ground-truth noise levels are known. But simulated objects are not easily able to replicate the underlying tissue attenuation heterogeneity, which will likely confound noise estimation methods. Therefore, we apply these approaches to patient image sets from low and high dose acquisitions.

This paper makes two key contributions. First, since absolute noise is not known, the expected relative change in noise content between two image sets corresponding to the same CT image scan is analyzed. A novel automated blind estimator metric is proposed to model the theoretical variations in CT image noise with variations in imaging dosages and techniques. Second, a performance optimization function is proposed to grade the proposed noise estimation methods in their ability to predict this known noise change between two studies. We evaluate the noise estimation performance of six viable filters in chest CT images.

## II. Method and Metrics

### A. Noise Estimation

The denoising performance of six different filtration methods was analyzed on sets of chest CT image data with varying image quality. The filters include the following:
1. Parameterized spatial approach of matched filtering (MF), where a known 2-D template with known variance is correlated with patches of CT images to estimate signal strength [6]
2. Spatial approach of bilateral filtering (BF), where the CT images are smoothened while preserving the edges. The variable filter parameters include filter size and estimated variance in two dimensions [2].

Manuscript received May 1, 2015. This work was supported in part by the NHLBI of the National Institutes of Health under grant number R01HL109327.

S, Roychowdhury and N. Hollcraft are with the *Division of STEM*, University of Washington Bothell (e-mail: roych@uw.edu).
A. Alessio is with the Department of Radiology at the University of Washington (e-mail: aalessio@uw.edu).

3. Parameterized 2-D anisotropic diffusion (AD), where the CT images are denoised while preserving the significant edges. The number of iterations, delta and kappa parameters can be varied for changing the filtering performance [7].
4. Spatial and frequency based approach of complex dual-tree wavelet transform (CDWT), where high frequency and low-frequency coefficients of each CT image is extracted in three stages. In each stage the low-frequency image is sub-sampled for the next stage of coefficient estimation. The threshold pixel value can be varied as a filter parameter [4].
5. Frequency based approach that minimizes a Fourier-domain error (FDE) for estimating the additive Gaussian noise variance in CT images using Wiener deconvolution. The noise variance estimate can be varied as a filter parameter [5].
6. Non-parameterized approach of Patch wise Non-local Means (PWNLM) that reduces noise in CT images using a patch-wise similarity indicator [3].

The residuals between the filtered image and original image provide a "noise" image. In other words, this follows the simple formula that the Original image = Filtered Image + Residual Image ($I_o = I_f + I_r$) [8]. If the filtered image primarily contains signal content, then the Original image = Signal Image + Noise Image as shown in Fig. 1. The variance within a pre-determined region of interest (ROI) in the noise image, $\sigma$, can be used as an estimate of additive noise.

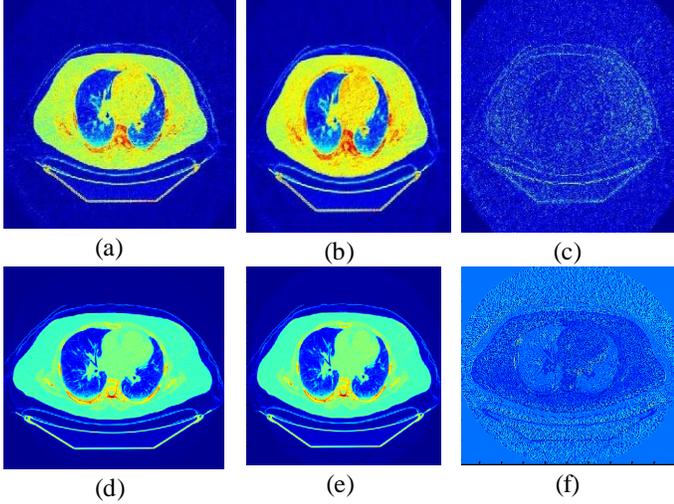

(a) (b) (c)
(d) (e) (f)

Fig 1. Example of residual noise image estimation using low and high dose techniques of CT imaging. (a)(b)(c) Correspond to low dose imaging technique, and (d)(e)(f) correspond to high dose imaging technique. (a)(d) Original image (b)(e) Filtered signal image (c)(f) Residual noise image.

### B. Patient Cases

In an IRB approved study, CT acquisitions were performed on two patients on a Lightspeed 16-slice scanner (GE Healthcare) with the clinical prescribed acquisition technique and then a lower dose acquisition technique over the identical scan range. Many scan settings were kept constant (helical, pitch 1.375, 20 mm collimation, 120kVp). In one patient, the high technique was performed with 48 mAs and the low with 5 mAs. In the other patient, the high technique was performed with tube current modulation with 190-280 mAs and the low with 5 mAs [9]. The ratio of anticipated variance change from the high to low study, which we denote as the theoretical ratio $R_{the}$, was calculated as the ratio of the tube current time product between the high and low dose study for each slice.

Noise estimation techniques discussed above were applied to each slice of the high and low dose exams. The ratio of the variance estimate from the high dose and estimate from the low dose study $R_{blind}$ provides an estimate of the noise change between the exams. $R_{blind}$ is estimated as the ratio between the noise variances in residual images from low and high dose CT scans within a ROI that represents the relatively uniform right para-spinal region as shown in (1).

$$R_{blind} = \frac{\sigma_l^2}{\sigma_h^2} \qquad (1)$$

### C. Metrics

If the noise estimation is successful, $R_{blind}$ should be the same as the theoretical $R_{the}$. In most noise-reduction filtering operations, increased filtration causes increased bias in the filtered image. This will lead to increased signal content in the residual image. To develop a method with an optimal tradeoff of noise and bias, we propose a cost function that can balance these effects:

$$\theta(\phi) = (1 - R_{blind}/R_{the})^2 + \beta M \qquad (2)$$

where, $M$ is the mean absolute error in the signal image defined as

$$M = \frac{1}{2}\left[\left|I_{o,1} - I_{f,1}\right| + \left|I_{o,2} - I_{f,2}\right|\right]. \qquad (3)$$

In (3), $I_{o,1}$ and $I_{f,1}$ the original and filtered image from the high technique and $I_{*,2}$ from the low technique. The hyper-parameter $\beta$ controls the relative strength of the absolute error and noise term. Our goal is to select the filter and filter parameters, $\phi$, that minimizes the cost function $\theta$.

### III. EXPERIMENTS AND RESULTS

Fig. 2 and Fig. 3 present examples of image slices from high and low dose acquisition techniques acquired from two different patients, respectively. In this work, since absolute noise levels are not known, we seek to select the method that is best able to predict the ratio of noise between these images. Corresponding to the two patients represented in Fig. 2 and Fig. 3, 12 chest CT scans are selected (6 images per patient) for the analysis of image noise content. Prior to image filtering, the estimates of $\frac{R_{blind}}{R_{the}}$ and $\theta$ for each of the 12 sets of CT scans are shown in Fig. 4. We observe that before filtering, for patient 1, $R_{blind}$ is under-estimated, while for patient 2 $R_{blind}$ is over-estimated.

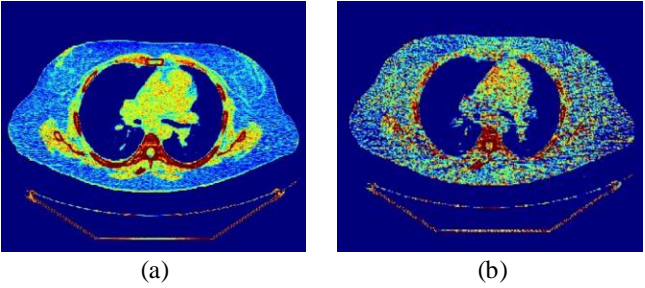

(a)                     (b)

Fig 2. Example of high and low dose slices from patient 1. (a) High dose image. (b) Low dose image. On this slice the the $R_{the}$ value is 7.36, meaning the tube current time product of the high dose study is 7.36 times higher than then low dose study. Assuming linear processing and image reconstruction, as performed here, the theoretical variance in the low dose study should be 1/7.36 times that in the high dose study.

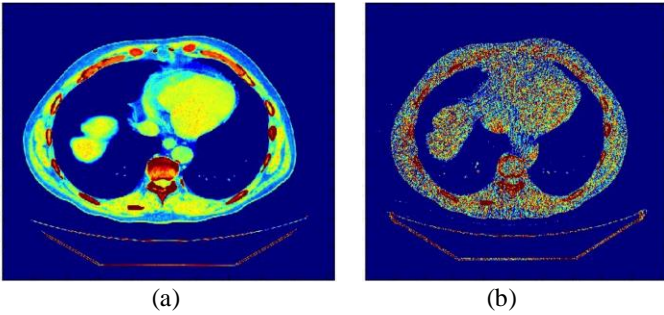

(a)                     (b)

Fig. 3. Example of high and low dose acquisitions from patient 2 in chest CT study. (a) High dose image. (b) Low dose image. On this slice, the $R_{the}$ value is 56.8, meaning the tube current time product of the high dose study is 56.8 times higher than then low dose study.

The matched filter (MF) and non-parametric (PWNLM) methods do not improve the $\frac{R_{blind}}{R_{the}}$ estimates significantly. The variation trends in $\frac{R_{blind}}{R_{the}}$ and $\theta$ as the filter parameters $\phi$ vary for the remaining 4 filters is shown in Fig. 5. For minimization of $\theta$ using (2), $\beta$ is varied as $[0.01, 0.1, 1, 10, 100, 1000]$.

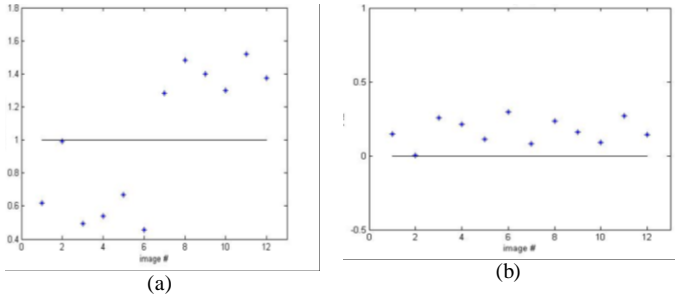

(a)                     (b)

Fig. 4. Theoretical vs automated blind estimators for noise variance before filtering. (a) $R_{blind} / R_{the}$, (b) $\theta$ estimates for 12 scans, respectively.

From the filtering trends in Fig. 5, the filter parameter that minimizes $\theta$ is chosen as optimal. Table 1 represents these optimal filter parameters ($\phi$) and the corresponding values of $\theta$ and $\beta$. It is noteworthy that for all the filtering methods, the $\theta$ vs. $\phi$ demonstrates similar trends for varying values of $\beta$.

TABLE 1: OPTIMAL FILTER PARAMETERS FOR MINIMIZING $\theta$

| Filter | Optimal $\phi$ | $\beta$ value | $\theta$ values |
|---|---|---|---|
| AD | Iteration = 20; delta = 8/40; kappa = 54 | 10 | 0.01285 |
| BF | window variance= [0.3 0.3], size[5x5] | 1 | 0.0912 |
| CDWT | Pixel-value Threshold = 150 | 10 | 0.01507 |
| FDE | Noise variance = 10E-9 | 1 | 0.0426 |

We observe that the best filtering operation using 2-D AD filter reduces the mean value of $\theta$ from 0.2249 to 0.012 while reducing mean $\frac{R_{blind}}{R_{the}}$ from 1.112 to 0.91. The FDE based filter shows a wide range of $\frac{R_{blind}}{R_{the}}$ and $\theta$ values with varying filter parameters. Thus, this Fourier-domain based filter is least robust for estimation of noise variance in CT images.

### IV. DISCUSSION AND CONCLUSION

In this work, we perform blind estimation of noise variance metrics from residual images. The goal is to blindly identify automatic image-specific metrics of noise that correlate with the variation in CT acquisition techniques. The blind estimator is optimized to correlate with theoretical prediction of CT image noise. We compare performance of several smoothing filters to select the filtering method that is best able to predict the ratio of noise between these images acquired by high and low techniques.

We observe that Frequency-based filtering methods (FDE) have the worst performance for blind estimation of CT image noise. On the other hand, combination of spatial and frequency domain based filtering approaches like AD and CDWT filters have the best blind estimation characteristics as shown in Fig. 6.

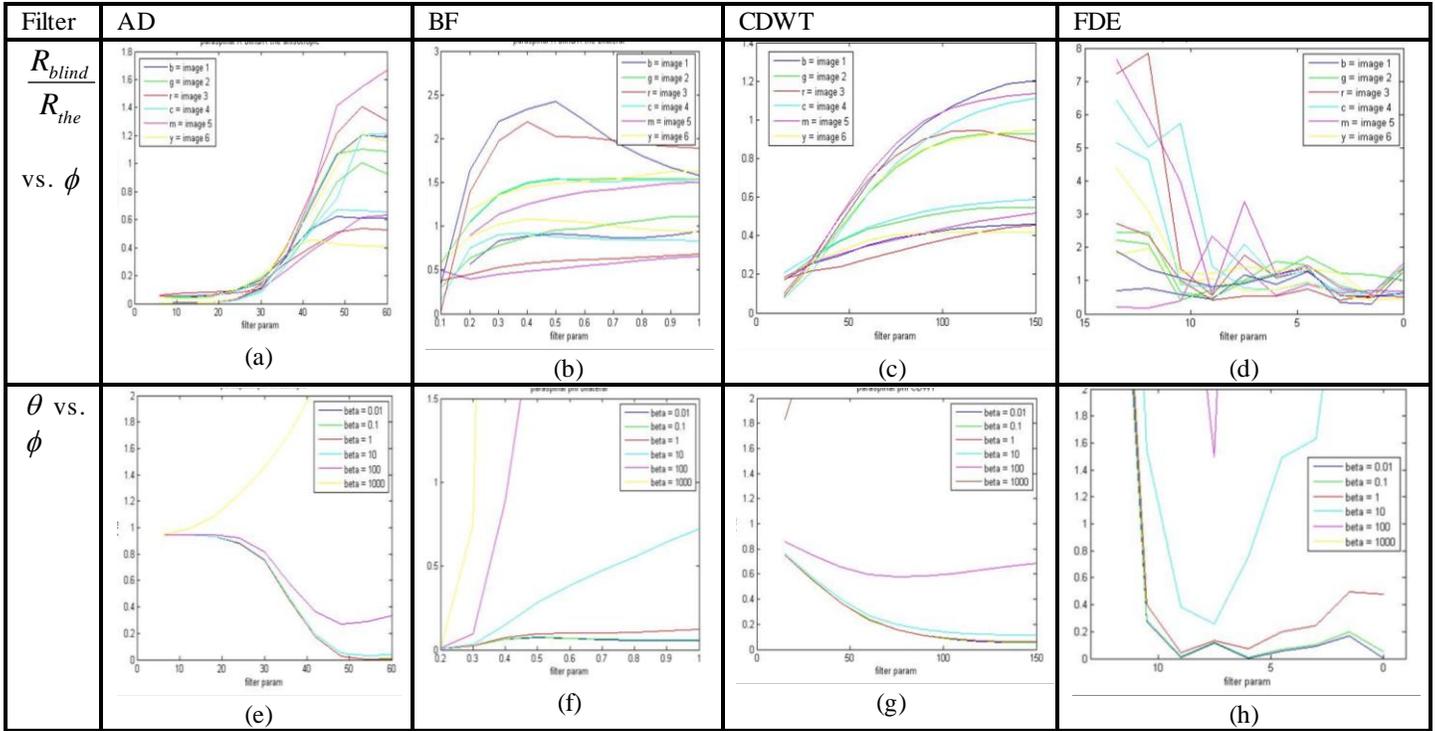

Fig. 5. $\frac{R_{blind}}{R_{the}}$ vs. $\phi$ results across for various filter parameters using (a) AD, (b) BF, (c) WS, (d) FDE based filters. $\theta$ vs. $\phi$ results across for various filter parameters using (e) AD, (f) BF, (g) WS, (h) FDE based filters.

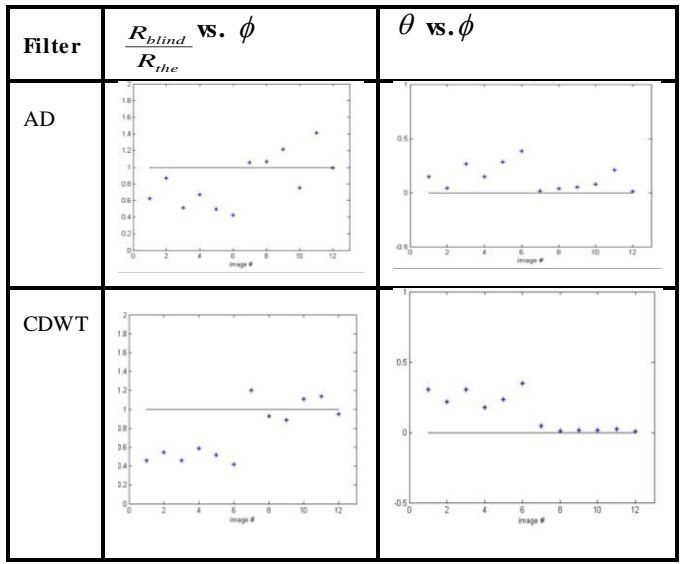

Fig. 6. Optimal filter performance for noise variance estimation.

Future efforts will be directed towards the assessment of noise variance in additional uniform regions in the chest and abdominal CT images. Analysis of noise variances in automatically extracted ROI masks followed by ranking of the automated metrics for multiple patients and multiple noise settings will further aid blind estimation of CT image noise content. Such automated blind estimators of noise content in CT images will be useful for protocol selection guidance in a quantitative, clinically feasible and sustainable manner by introducing automated feedback regarding image quality for low radiation dose images.